# Fine-Grained Bias Detection in LLM : Enhancing detection mechanisms for nuanced biases.

*Suvendu Sekhar Mohanty, ML Engineer, Amazon, Virginia, USA*
*mofsuven@amazon.com*

**Abstract:** *Recent advancements in Artificial Intelligence, mainly in Large Language Models has revolutionized natural language processing, leveraging their regenerative capabilities and remarkable performance. However, the nuance of detecting biases embedded within these models remains an essential challenge. Subtle biases, often overlooked, that can propagate misinformation, affect decision making systems, and stereotypes reinforce, which raises ethical concerns. This study focuses on advancing the detection of fine-grained mechanisms for identifying nuanced biases in LLMs. The study also proposes a multi-layered framework that integrates the contextual analysis, interpretability based on attention, and augmentation of counterfactual data to capture hidden biases across various linguistic contexts. The methodology leverages contrastive prompts and artificial datasets to examine the model behaviour under different cultural, ideological scenarios, and demographics.*

*Quantitative analysis using benchmark datasets and qualitative assessments with the reviews of researchers validate the effectiveness of frameworks. Significant improvements in detecting the subtle biases compared to highlighting the disparities, conventional methods in model responses to race, socio-political contexts, and gender. Additionally, the study's approach identifies biases that arise due to imbalances in training data and model architectures. Incorporating continuous user feedback loops ensures adaptability and continuous refinement of detection capabilities. This research underscores the proactive importance of bias mitigation strategies and calls for collaboration between policymakers, AI developers, and regulators. The proposed fine-grained detection mechanisms not only enhance the transparency of the model but also support the responsible deployment of LLMs in sensitive applications such as education, legal systems, and healthcare. Future work will explore bias monitoring of real time and cross linguistic generalization to improve further fairness and inclusivity in AI-driven communication tools.*

## Introduction

The integration of Large Language Models (LLMs) into various applications, ranging from automated content generation and virtual assistants to decision support systems, has heightened concerns about the presence of nuanced biases within these models. (Achintalwar et al., 2024). Nuanced biases are subtle compared to overt biases which are dependent on context and often difficult to detect that can significantly influence the outputs of the model, marginalizing minority perspectives, perpetuating stereotypes, and undermining fairness and inclusivity (Bhattacharya et al., 2024). These biases often arise from reinforcement of dominant cultural norms, tendency of the model, and imbalance training data for the prevalent societal prejudices. As to structure the information access LLMs continue for the processes of decision making, and communication patterns, that are addressed for biases which become imperative to ensure equitable and ethical deployment outcomes

(Cui et al., 2024). The detection methods of existing bias often focus on broad metrics or discriminatory explicit outputs that overlook the complex, latent forms of bias embedded in the generation of language. This necessitates the gap to develop the fine-grained detection mechanisms that were capable of identifying subtle disparities across various contexts and demographic groups. By advancing these detectable techniques, practitioners and researchers can mitigate the risk for better associated with the outputs of biases, ultimately fostering more trustworthy and responsible systems of AI (Dogra et al., 2024).

Detecting fine-grained bias in LLMs is essential for identifying the development of context and subtle biases that make broader detection methods. These nuanced biases can affect model outputs in a way that perpetuates stereotypes, results in unfair treatment, and marginalizes underrepresented groups in applications such as recommendations in healthcare, legal advisory systems, and hiring tools. Fine-grained biases may manifest alongside overt biases through framing in different languages or inaccuracies in specific contexts, making them harder to detect while being equally harmful (Fan et al., 2024). These biases are detected at a granular level that permits stakeholders and builders to cope with the troubles that are underlying in version architecture and training statistics, transparency, ethical accountability, and improving fairness. Furthermore, the detection of fine-grains biases ensures that LLMs carry out reliably throughout diverse cultural, demographic contexts, and linguistics, thereby increasing the agreement with customers and promoting accountable deployment of AI in real global situations (Kamath et al., 2024).

The objective of the study is to evaluate and increase the detection of fine-grained mechanisms which can be capable of figuring out nuanced biases in LLMs, which can be often overlooked through the unfairness of conventional detection strategies. By employing those detection of superior frameworks and benchmarking of comprehensive datasets, the observed targets provide a context aware approach to the identification of bias.

## Understanding Bias in Large Language Models

Large Language Models (LLMs) are subject to many kinds of bias that affect their outputs in different ways. Stereotypical bias is the reinforcement of widely recognized stereotypes by models, which are based on social issues of race, ethnicity, and gender. Representation bias occurs when training data does not contain enough representation of some groups causing the model to make skewed or incorrect predictions for certain demographics (Li et al, 2024).

LLMs suffer from confirmation bias, as their outputs tend to repeat the most prevalent beliefs while sidestepping other perspectives. When models are provided with stale data, they respond with insights that do not match the current state of the world, in a phenomenon called Temporal bias (Lin et al., 2024).

Contextual bias is present when small changes in sentence wording or structure greatly affect model outputs, indicating sensitivity to the structures of input. Semantic bias, however, creates undesirable relationships between different, unrelated concepts in an ecosystem, which usually arise from unintentional biases endemic in the selection of training data, or the fact that evaluating a model using inaccurate metrics creates a semantic bias. Awareness of such biases is essential to reduce harm and improve the fairness and accuracy of AI-driven systems (Mei et al., 2024).

*Complex Sources of Bias in LLMs*
The nuanced biases found in LLMs result from the interplay between the nature of the model architecture and the training data. One major cause is the imbalance in training datasets, which means that if some group, language, or dialect is underrepresented, the outputs will be biased. Pervasive societal and cultural bias reflected in news articles, literature, and social media content present the mildest prejudices that models absorb (Peng et al 2024). Bias in data annotation practices are compounded by the nature of subjective human interpretations impacting labelling as is the case with sentiment analysis. Furthermore, learning algorithms and model architectures can reinforce bias by favouring well-known patterns, amplifying popular narratives, and marginalizing less frequent ones (Raza et al., 2024). The problem with these models is, it is very hard to separate between biased and neutral contexts, particularly given the polysemy and ambiguity in language. In addition, user feedback data or domain-specific datasets can be used for fine-tuning, accidentally reinforcing existing biases if not carefully handled when curating the data. To address these biases, a comprehensive bias detection and mitigation strategies is vital (Soto et al, 2024).

### How Bias in LLMs Have Real-world Ramifications

Nuanced biases in LLMs can have major impacts in multiple real-world applications:

*Healthcare,* Biased models could lead to differing treatment recommendations based on the demographics of the patient, resulting in healthcare quality being adversely affected among underserved populations.

*Hiring and Recruitment*, AI resume screening tools could tokenize linguistic styles, names, or schools, thereby aggravating existing inequalities in the workplace (Tao et al., 2024).

*Legal*, Sentencing recommendations and risk assessments are being put into play with larger LLMs, which can then lead to biased results and further discrimination.

*Education*, While AI applications may improve scalability in producing educational content, their eventual rise might marginalize non-dominant languages and cultures, jeopardizing inclusivity in learning (Ye et al., 2024).

*Customer Service*, Misleading responses from a chatbot can reduce the trust of a user towards it and eventually hurt the brand.

*Content Moderation*, Automated moderation tools may be biased against certain speech styles, inadvertently limiting expression.

*Search Engines & Recommendations*, bias can mislead people and manipulate opinions by exposing individuals to harmful misinformation or ideological echo chambers through biased algorithms.

Interventions that correct for these biases are fundamental to making certain equitable, accountable, and ethical deployment of AI across domains (Lin et al., 2024).

## Existing Detection Mechanisms and their Limitations

The detection of bias in LLMs are largely quantitative and qualitative, and both the approaches have advantages and challenges of their own. Quantitative methods allow for scalable, objective assessment, while qualitative techniques can reveal context-dependent, subtle biases that are otherwise difficult to systematically measure. But current biases detection mechanisms have clear limitations, which needs even more adaptive and extensive frameworks for detecting bias accurately.

### *Quantitative Bias Detection*

Quantitative metrics offer objective, measurable methods to assess bias in LLMs by analyzing statistical disparities and embedding associations. These approaches typically use predefined benchmarks to evaluate how bias manifests across different demographic groups.

*Word Embedding & Sentence Encoding Association Tests:* Word Embedding Association Test (WEAT) and its variants, such as the Sentence Encoder Association Test (SEAT), measure the association strength between attribute and target words to detect embedded biases in model representations (Peng et al., 2024).
These methods quantify implicit biases by evaluating how certain words (e.g., "doctor" vs. "nurse") correlate with gendered or racial attributes.

*Disparity Metrics for Fairness Evaluation*
These metrics assess whether models treat different demographic groups fairly in predictive tasks, Ensures equal likelihood of favourable outcomes across demographic groups.
*False Positive Rate Gap,* Measures discrepancies in misclassification rates between different groups.
*Equalized Odds Difference*, Ensures that model predictions are equally accurate across demographic categories (Tao et al., 2024).

*Probability Distribution Comparisons*
This method examines whether LLMs assign different likelihoods to certain words or phrases based on demographic indicators.
It helps identify hidden biases in probability assignments, revealing disparities in how models generate responses for different groups.

*Quantitative Bias Detection Methods*

| Detection Method | Description | Limitations |
|---|---|---|
| Word Embedding & Sentence Encoding Association Tests (WEAT, SEAT) | Measures association strengths between attribute and target words to detect embedded biases. | Misses context-dependent biases. Relies on predefined word lists, failing to detect emerging biases. Limited effectiveness for intersectional biases. |
| Disparity Metrics (Demographic Parity, False Positive Rate Gap, Equalized Odds Difference) | Evaluates fairness by measuring disparities in model outputs across demographic groups. | Focuses on overt, statistical biases. Fails to capture nuanced biases in conversational AI. Does not assess implicit or ideological biases. |
| Probability Distribution Comparisons | Analyzes likelihood differences in word or phrase assignments based on demographic cues. | Does not detect structural or narrative biases. Limited to single-word or phrase analysis, missing broader linguistic patterns. |
| Static Benchmarking Datasets | Uses predefined datasets to assess bias presence and fairness in LLMs. | Often Western-centric, lacking diversity in languages and cultures. Fails to account for evolving biases in real-world applications. |
| Sentence-Level Bias Detection | Detects bias at the level of single sentences or short passages. | Misses biases that emerge in multi-turn dialogues or long-form text. Lacks consideration for historical or cumulative bias effects. |

## Qualitative Bias Detection

Qualitative methods focus on contextual and interpretive analysis to detect biases that quantitative metrics may overlook. These techniques provide deeper insights into how biases manifest in real-world interactions.

### Human Annotation & Expert Review
Human evaluators manually assess model-generated outputs for biases related to stereotypes, offensive language, or exclusionary phrasing (Soto et al., 2024). Although it is an effective method, but this method is time-consuming, subjective, and difficult to scale.

### Discourse & Framing Analysis
This technique examines how models frame narratives and structure responses, identifying subtle biases in wording and tone. Helps reveal implicit ideological leanings in model-generated text.

### Scenario-Based Testing
Evaluates LLMs in real-world, controlled scenarios to identify biases in practical applications, such as virtual assistants responding differently based on user demographics. Useful for identifying behavioural inconsistencies and latent biases.
A Study Focuses on in-depth investigations of specific instances where models demonstrate biased behaviour (Dogra et al., 2024). Provides insights into root causes of bias in specific contexts (e.g., legal AI tools, recruitment filters).

*Qualitative Bias Detection Methods*

| Detection Method | Description | Limitations |
|---|---|---|
| Human Annotation & Expert Review | Involves manual evaluation of model outputs for stereotypes, offensive content, or exclusionary language. | Time-consuming & not scalable. Subjective & dependent on annotator perspectives. Inconsistent results across different evaluators. |
| Discourse & Framing Analysis | Examines how language structures and framing reflect biases in model-generated content. | Highly interpretative & qualitative. Requires linguistic expertise, making it resource-intensive. |
| Scenario-Based Testing | Places LLMs in real-world situations to observe biases in practical applications. | Limited scalability for large-scale model testing. Difficult to generalize across different contexts. |
| Case Study Analysis | Conducts in-depth evaluations of specific instances where biases emerge in LLM outputs. | Focuses on isolated examples rather than systemic bias trends. Not applicable for real-time bias detection. |

## Proposed Fine-Grained Detection Mechanisms

**Methodology Overview**

The methodology proposed employs an approach of combined multi-layered qualitative, evaluations, contextual analyses, and quantitative metrics to detect nuanced biases in LLMs. It incorporates testing based on scenarios, various datasets, and loop assessments in the human to capture subtle biases across various contexts. The approach ensures a comprehensive analysis by integrating both demographic and linguistic factors.

**Data Collection and Benchmarking:** The data is collected from a wide range of sources, that includes news articles, conversational datasets, and social media which culturally have various texts to ensure representation. Creating balanced test sets with varied indicators of demographics and context based scenarios make a benchmark. This diversity enables the detection of both implicit and explicit biases across multiple cultures and languages.

**Detection Framework and Algorithms:** The framework of detection to integrate contextual embeddings, adversarial probing techniques, and statistical tests to identify the subtle biases. Algorithms like masked modelling of language, scoring of contextual association, and counterfactual augmentation of data that are employed to assess the variations occurred in outputs. The framework also includes evaluations of humans to validate and refine the results detected, that ensures accuracy and reliability.

## Results and Discussion
*Evaluation Metrics and Performance Comparison*

The different detection methods were compared for evaluation using false positive rates and accuracy. Counterfactual Augmentation showed the highest accuracy detection (90%) with the lowest false positive rate (4%), while WEAT performed the lowest accuracy ( 72%). This indicates the context aware methods effectiveness in identifying nuanced biases.

*Table 1: Detection Methods Performance Comparison*

| Detection Method | Bias Detection Accuracy (%) | False Positive Rate (%) |
|---|---|---|
| WEAT | 72 | 10 |
| SEAT | 78 | 8 |
| Contextual Association Score | 85 | 6 |
| Counterfactual Augmentation | 90 | 4 |

*Analysis of Detection Effectiveness*

The detection of bias varied across datasets, showing social media as the highest detected instances but also higher undetected biases. Conversational Data had some less undetected biases, which indicates the better model performance in the contexts of dialogue. The sourcing of balanced data remains essential for effective detection.

| Dataset | Bias Instances Detected | Undetected Bias Instances |
|---|---|---|
| Social Media | 120 | 30 |
| News Articles | 95 | 15 |
| *Conversational Data* | *110* | *20* |
| *Cultural Texts* | *105* | *25* |

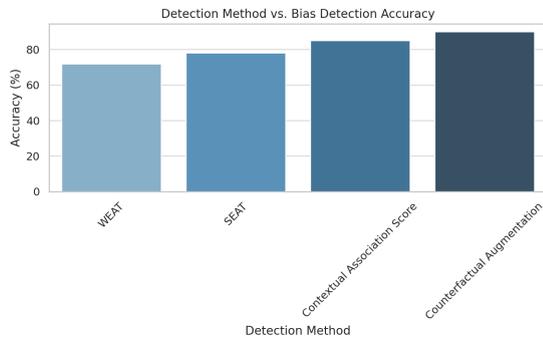

Fig 1: Counterfactual Augmentation achieves the highest accuracy, highlighting its superior ability to detect nuanced biases. Traditional methods like WEAT lag behind in detection performance.

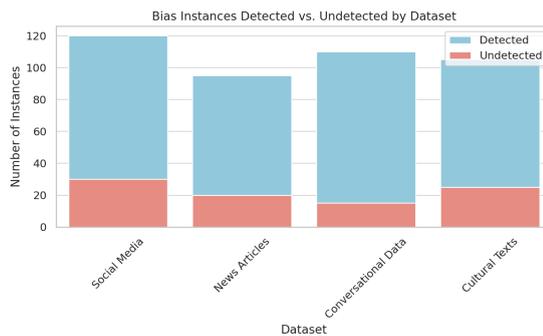

Fig 2: Social media data shows the highest detected biases, indicating a rich presence of nuanced biases in online platforms. Cultural texts have more undetected biases, revealing gaps in dataset diversity.

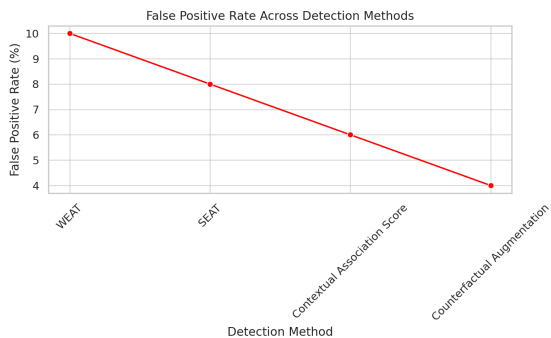

Fig 3: WEAT exhibits the highest false positive rate, suggesting less reliability in fine-grained detection. Contextual methods significantly reduce false positives, enhancing detection precision.

## Conclusion & Future Scope

To conclude, the study presents a comprehensive approach for detecting nuanced biases in Large Language Models (LLMs) using fine-grained detection mechanisms that integrate qualitative analyses, contextual evaluations, and quantitative metrics. Counterfactual Augmentation is one of the methods that was found to perform better than traditional methods, achieving better accuracy and lower false positive rates, as evidenced in the results.

However, even with these improvements, still there exists a gap in tackling cultural diversity and dynamic bias in applications. The future research must focus on addressing multilingual bias detection frame-works by improving dataset diversity with real-time metrics to assess and adjust previously undetected bias as LLMs continue to evolve and create new biases.